\documentclass[10pt,twocolumn,letterpaper]{article}

\usepackage{xcolor}

\usepackage[pagenumbers]{cvis} 

\definecolor{cvisblue}{rgb}{0.21,0.49,0.74}
\usepackage[pagebackref,breaklinks,colorlinks,allcolors=cvisblue]{hyperref}

\def\paperID{10} 
\def\confName{CVIS}
\def\confYear{2026}

\title{PortionNet: Distilling 3D Geometric Knowledge for Food Nutrition Estimation}

\author{
Darrin Bright \\
Vellore Institute of Technology \\
{\tt\small darrin.bright2022@vitstudent.ac.in}
\and
Rakshith Raj \\
Vellore Institute of Technology \\
{\tt\small rakshith.raj2022@vitstudent.ac.in}
\and
Kanchan Keisham \\
Vellore Institute of Technology \\
{\tt\small kanchan.keisham@vit.ac.in}
}

\begin{document}
\maketitle
\begin{abstract}
\label{sec:abstract}

Accurate food nutrition estimation from single images is challenging due to the loss of 3D information. While depth-based methods provide reliable geometry, they remain inaccessible on most smartphones because of depth-sensor requirements. To overcome this challenge, we propose PortionNet, a novel cross-modal knowledge distillation framework that learns geometric features from point clouds during training while requiring only RGB images at inference. Our approach employs a dual-mode training strategy where a lightweight adapter network mimics point cloud representations, enabling pseudo-3D reasoning without any specialized hardware requirements. PortionNet achieves state-of-the-art performance on MetaFood3D, outperforming all previous methods in both volume and energy estimation. Cross-dataset evaluation on SimpleFood45 further demonstrates strong generalization in energy estimation.
\end{abstract}    
\section{Introduction}
\label{sec:intro}

Dietary monitoring plays an important role in managing chronic diseases like obesity and diabetes, yet manual food logging suffers from high error rates~\cite{chotwanvirat2024advancements}. Image-based methods offer a solution to replace manual logging, but accurately estimating nutrition from a single RGB image is difficult due to the absence of geometric features. 

Depth sensor methods address this issue by providing reliable 3D geometry~\cite{dehais2017two}, but these methods are limited by the cost and accessibility of hardware requirements like LiDAR on smartphones. Therefore, RGB-only methods that do not require specialized hardware can be utilized on smartphones. However, these methods have difficulties with spatial reasoning, as current state-of-the-art RGB-only methods exhibit a mean absolute percentage error (MAPE) of 68.05\%~\cite{ma2024mfp3d}. This significant accuracy gap has 
prevented RGB-only methods from practical deployment.

Recent advances in knowledge distillation~\cite{hinton2015distilling,gupta2016cross} enable cross-modal capability transfer. However, existing food estimation methods either 
require depth sensors during inference or fail to utilize 3D effectively. No prior work has explored distilling 3D 
spatial reasoning into RGB-only networks for food nutrition estimation.

To address this gap, we propose PortionNet, a novel cross-modal knowledge 
distillation framework that learns 3D-aware representations from point 
clouds during training but requires only RGB during inference. A Point Cloud 
Geometry Encoder serves as a teacher, while a lightweight RGB-to-Geometry 
Adapter learns to mimic its feature space. Dual-mode training alternates 
between multimodal (RGB + point cloud) and RGB-only batches, limiting over-reliance on 3D information while encouraging strong pseudo-3D feature learning.

Our contributions can be summarized as follows:

\begin{itemize}[leftmargin=*,noitemsep]
    \item We propose PortionNet, a novel cross-modal knowledge distillation framework that allows RGB models to learn 3D geometric features from point clouds during training, eliminating depth sensor requirements at inference.
    
    \item We introduce a dual-mode training strategy with a lightweight RGB-to-Geometry Adapter that learns to generate pseudo-3D features for accurate geometric reasoning from standard RGB images.
    
    \item Our method achieves state-of-the-art performance on MetaFood3D~\cite{chen2024metafood3d} with 17.43\% volume MAPE and 15.36\% energy MAPE. Cross-dataset evaluation on SimpleFood45~\cite{vinod2024food} achieves state-of-the-art energy estimation (12.17\% MAPE), demonstrating effective generalization.
\end{itemize}

\section{Related Work}
\label{sec:related}

Early food portion estimation systems relied heavily on geometric primitives and manual calibration~\cite{fang2015single, yue2012measurement}, limiting their applicability to real-world scenarios. RGB-D methods~\cite{dehais2017two, konstantakopoulos2021reconstruction, kim2023food, lo2019food} substantially improved accuracy by utilizing explicit depth information for 3D reconstruction. However, these methods require specialized depth sensors such as LiDAR or structured light cameras, which remain unavailable on most consumer smartphones.

Modern RGB-only methods attempt to infer 3D information from 2D images through various strategies. Transformer-based methods~\cite{sheng2022lightweight, feng2023fine} use self-attention mechanisms for fine-grained food recognition, while depth prediction approaches~\cite{shao2023dpf} synthesize pseudo-depth maps to guide volume estimation. Multi-scale networks~\cite{wang2024visual} combine segmentation and regression for nutrient prediction. Lately, 3D model-based methods~\cite{vinod2024food} use existing 3D food models for portion estimation from single images. Despite these advances, RGB-only methods face significant challenges in recovering 3D geometry from monocular images, where scale information is fundamentally lost. MFP3D~\cite{ma2024mfp3d}, the current state-of-the-art RGB-only method, achieves 41.43\% volume MAPE and 68.05\% energy MAPE.

Knowledge distillation~\cite{hinton2015distilling} enables student models to mimic stronger teachers by matching their learned representations. Cross-modal distillation~\cite{gupta2016cross} extends this pattern to transfer knowledge between different input modalities, such as from depth to RGB. Recent work explores bidirectional learning~\cite{huo2024c2kd}, attention-based feature alignment~\cite{wang2024multiscale}, and handling incomplete modalities~\cite{zhang2024cross, kwak2025cross}, where certain sensors are unavailable at inference. However, these methods typically focus on tasks where semantic understanding is enough. Food nutrition estimation uniquely requires both semantic recognition (food type and composition) and metric geometric reasoning (absolute volume). To our knowledge, no prior work has explored distilling 3D spatial reasoning from point clouds into RGB-only networks for this task.
\section{Methodology}
\label{sec:method}

\subsection{Architecture Overview}
\begin{figure*}[t]
\centering
\includegraphics[width=\textwidth]{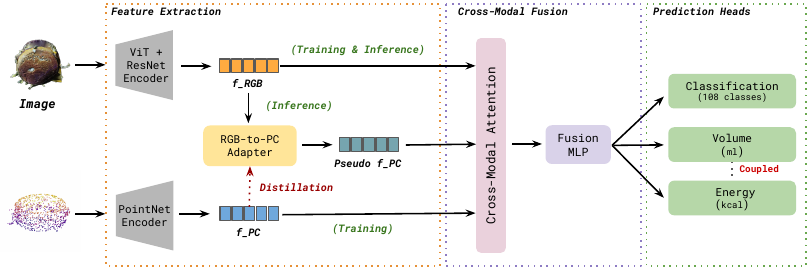}
\caption{Overview of the proposed PortionNet framework.}
\label{fig:architecture}
\end{figure*}

PortionNet learns 3D geometric representations by using point-cloud features during training while requiring only RGB images at inference. As shown in Figure~\ref{fig:architecture}, the framework consists of: (1) dual RGB encoders that extract visual features, (2) an augmented PointNet geometry encoder that serves as the teacher, (3) an RGB-to-point-cloud adapter that maps RGB features to geometry-aware space, and (4) a fusion module with task-specific prediction heads.

\subsection{Multi-Modal Feature Extraction}

\subsubsection{Dual RGB Encoders}
We extract visual features using pretrained ViT-B/16 and ResNet-18 encoders, concatenating their outputs (1280-d) before projecting to 256 dimensions via a two-layer MLP, resulting in feature vectors of shape [B × 256]. 

\subsubsection{PointNet Geometry Encoder}
We use PointNet~\cite{qi2017pointnet} to extract geometric features from point clouds, which learns spatial structure from unordered 3D data. We make two modifications: (1) append learned embeddings of bounding box dimensions ($\mathbb{R}^3 \to \mathbb{R}^{64}$) for explicit scale information, and (2) apply multi-scale adaptive pooling at {64, 128, 256, 512} resolutions before global aggregation. The final representation is projected to 256 dimensions.

\subsubsection{RGB-to-Point-Cloud Adapter}
The adapter is a lightweight two-layer MLP that maps the extracted RGB features into a geometry-aware embedding space. This adapter enables RGB-only inference by mimicking PointNet's output, eliminating the need for sensors while preserving geometric reasoning.

\subsection{Cross-Modal Knowledge Distillation}

\subsubsection{Dual-Mode Training}
Training alternates between multimodal mode using both RGB and PointNet features (70\%) and RGB-only mode using RGB and adapter-generated features (30\%), preventing over-reliance on real point clouds while ensuring strong pseudo-3D learning.

\subsubsection{Distillation Objective}
We supervise the adapter to mimic PointNet's geometric features. Let $f_{\text{adapter}} \in \mathbb{R}^{256}$ and $f_{\text{PointNet}} \in \mathbb{R}^{256}$ denote the feature embeddings produced by the RGB adapter and the PointNet encoder, respectively. The distillation loss combines three complementary objectives:
\begin{equation}
\label{eq:distill_total}
\mathcal{L}_{\text{distill}}
= w_{\text{mse}}\,\mathcal{L}_{\text{MSE}}
+ w_{\text{cos}}\,\mathcal{L}_{\text{cos}}
+ w_{\text{KL}}\,\mathcal{L}_{\text{KL}}
\end{equation}
where $w_{\text{mse}} = 0.7$, $w_{\text{cos}} = 0.2$, $w_{\text{KL}} = 0.1$. The MSE loss
\begin{equation}
\label{eq:mse_loss}
\mathcal{L}_{\text{MSE}} = \lVert f_{\text{adapter}} - f_{\text{PointNet}} \rVert_2^2
\end{equation}
matches feature magnitudes, the cosine loss
\begin{equation}
\label{eq:cos_loss}
\mathcal{L}_{\text{cos}} = 1 - \cos(f_{\text{adapter}}, f_{\text{PointNet}})
\end{equation}
aligns feature directions, and the KL divergence loss
\begin{equation}
\label{eq:kl_loss}
\mathcal{L}_{\text{KL}} = \mathrm{KL}\!\left(\sigma(f_{\text{PointNet}}/T) \,\|\, \sigma(f_{\text{adapter}}/T)\right)
\end{equation}
matches probability distributions with softmax $\sigma$ and temperature $T=4$.

\subsection{Feature Fusion and Prediction}

\subsubsection{Cross-Modal Fusion}
RGB and point cloud features are fused using bidirectional cross-attention with residual connections and layer normalization. The attended features are concatenated and processed through a fusion MLP.

\subsubsection{Task-Specific Prediction Heads}
We employ three prediction heads: (1) 3-layer MLP for the 108-class food classification, (2) volume estimation with Softplus activation, ensuring 
positive outputs, and (3) energy estimation using concatenated features to model physical volume-energy dependency.

\subsection{Training Objective}
The overall training objective combines three weighted loss components:
\begin{equation}
\label{eq:total_loss}
\mathcal{L}_{\text{total}} = \lambda_{\text{cls}}\mathcal{L}_{\text{cls}} + \lambda_{\text{reg}}\mathcal{L}_{\text{reg}} + \lambda_{\text{distill}}\mathcal{L}_{\text{distill}}
\end{equation}
where $\mathcal{L}_{\text{cls}}$ is the classification loss, $\mathcal{L}_{\text{reg}}$ is the regression loss for volume and energy prediction, $\mathcal{L}_{\text{distill}}$ is the feature distillation loss (Eq.~\ref{eq:distill_total}), and $\lambda_{\text{cls}}$, $\lambda_{\text{reg}}$, $\lambda_{\text{distill}}$ are their respective task weights.

\noindent\textbf{Classification Loss.} We employ cross-entropy with label smoothing ($\epsilon = 0.05$):
\begin{equation}
\label{eq:cls_loss}
\mathcal{L}_{\text{cls}} = -\sum_{i=1}^{C} y_i \log(\hat{y}_i)
\end{equation}
where $C=108$, $y_i$ is the smoothed label, and $\hat{y}_i$ is predicted probability.

\noindent\textbf{Regression Loss.} We employ a combination of L1 and Huber losses ($\delta = 0.5$) for both volume and energy prediction. Let $v$ and $\hat{v}$ denote the ground-truth and predicted food volume, and $e$ and $\hat{e}$ denote the ground-truth and predicted energy values. The regression loss is calculated as:
\begin{equation}
\label{eq:reg_loss}
\begin{aligned}
\mathcal{L}_{\text{reg}} =\;& 0.4 \big( \text{L1}(v, \hat{v})
+ \text{Huber}(v, \hat{v}, \delta=0.5) \big) \\
&+ 0.6 \big( \text{L1}(e, \hat{e})
+ \text{Huber}(e, \hat{e}, \delta=0.5) \big)
\end{aligned}
\end{equation}
where $\hat{e}$ is batch-normalized for scale invariance.

\subsection{Implementation Details}
We train PortionNet using the AdamW optimizer with differential learning rates ($1 \times 10^{-4}$ for pretrained encoders, $5 \times 10^{-4}$ for heads 
and adapter), OneCycleLR scheduling (10\% warmup, 25 epochs), an effective batch size of 64 (16 $\times$ 4 accumulation), 8-head cross-modal attention, and gradient clipping (norm 1.0). We employ GradNorm~\cite{chen2018gradnorm} with 
$\lambda_{\text{reg}} = 0.1$ and $\lambda_{\text{distill}} = 0.5$ in multimodal mode.

\section{Results}
\label{sec:results}

We evaluate PortionNet on MetaFood3D~\cite{chen2024metafood3d}, containing 
637 objects across 108 classes with 12,752 RGB images and point clouds. We follow MFP3D's 80/20 split. For cross-dataset evaluation, we use SimpleFood45~\cite{vinod2024food} with 513 images across 12 classes. All results are averaged over 3 seeds with standard deviations.

\subsection{Evaluation on MetaFood3D}

Table~\ref{tab:metafood_modes} shows that PortionNet achieves 98.34\% classification accuracy and strong regression (17.43\% volume MAPE and 15.36\% energy MAPE) using only standard RGB cameras for inference. The minimal gap between RGB-only and RGB+PC modes validates effective knowledge distillation.

\begin{table}[h]
\centering
\caption{MetaFood3D Results: RGB-only vs RGB+PC modes} 
\label{tab:metafood_modes}
\begin{tabular}{lcc}
\toprule
\textbf{Metric} & \textbf{RGB} & \textbf{RGB+PC} \\
\midrule
R$^2$ & 0.926 $\pm$ 0.004 & \textbf{0.926 $\pm$ 0.005} \\
Accuracy (\%) & \textbf{98.34 $\pm$ 0.58} & 98.26 $\pm$ 0.60 \\
Vol. MAE (mL) & 25.39 $\pm$ 2.59 & \textbf{25.16 $\pm$ 2.41} \\
Vol. MAPE (\%) & 17.43 $\pm$ 0.81 & \textbf{17.33 $\pm$ 0.84} \\
Eng. MAE (kcal) & \textbf{32.26 $\pm$ 0.43} & 32.64 $\pm$ 0.66 \\
Eng. MAPE (\%) & \textbf{15.36 $\pm$ 1.33} & 15.39 $\pm$ 1.46 \\
\bottomrule
\end{tabular}
\end{table}

\subsection{Comparison with RGB-based Methods}

Table~\ref{tab:rgb_comparison} shows that PortionNet achieves substantial improvements over the state-of-the-art RGB-only method MFP3D~\cite{ma2024mfp3d}, reducing volume MAPE from 41.43\% to 17.43\% (57.9\% relative improvement) and energy MAPE from 68.05\% to 15.36\% (77.4\% relative improvement), demonstrating effective knowledge distillation without requiring depth sensors at inference.

\begin{table}[h]
\centering
\caption{Comparison with other methods on MetaFood3D}
\label{tab:rgb_comparison}
\resizebox{\columnwidth}{!}{%
\begin{tabular}{lcc}
\toprule
\textbf{Method} & \textbf{Vol. MAPE (\%)}~\textcolor{blue}{$\downarrow$} & \textbf{Eng. MAPE (\%)}~\textcolor{blue}{$\downarrow$} \\
\midrule
Density Map~\cite{shao2021towards}$^{\dagger}$ & -- & 663.43 \\
Stereo~\cite{dehais2017two}$^{\dagger}$ & 210.90 & -- \\
Voxel~\cite{shao2023end}$^{\dagger}$ & 104.07 & -- \\
3D Assisted~\cite{vinod2024food}$^{\dagger}$ & 79.33 & 102.25 \\
MFP3D~\cite{ma2024mfp3d}$^{\dagger}$ & 41.43 & 68.05 \\
\midrule
\textbf{PortionNet (Ours)} & \textbf{17.43 $\pm$ 0.81} & \textbf{15.36 $\pm$ 1.33} \\
& \textcolor{blue}{(-24.0)} & \textcolor{blue}{(-52.7)} \\
\bottomrule
\end{tabular}%
}
\vspace{0.1cm}
\footnotesize{$^{\dagger}$Results from MFP3D~\cite{ma2024mfp3d}}\\
\end{table}
\subsection{Cross-Dataset Generalization}

We also perform cross-data generalization on the SimpleFood45 dataset as shown in Table~\ref{tab:cross_dataset}. PortionNet achieves 12.17\% energy MAPE, a 49.4\% relative improvement over MFP3D~\cite{ma2024mfp3d} (24.03\%), demonstrating strong cross-dataset generalization.

\begin{table}[h]
\centering
\caption{Cross-dataset generalization on SimpleFood45}
\label{tab:cross_dataset}
\resizebox{\columnwidth}{!}{%
\begin{tabular}{lcc}
\toprule
\textbf{Method} & \textbf{Vol. MAPE (\%)}~\textcolor{blue}{$\downarrow$} & \textbf{Eng. MAPE (\%)}~\textcolor{blue}{$\downarrow$} \\
\midrule
Density Map Only$^{\dagger}$ & -- & 159.48 \\
Density Map Summing$^{\dagger}$ & -- & 93.16 \\
Voxel Reconstruction$^{\dagger}$ & 24.51 & -- \\
3D Assisted$^{\dagger}$ & 14.01 & 25.13 \\
MFP3D$^{\dagger}$ & 16.15 & 24.03 \\
\midrule
\textbf{PortionNet (Ours)} & \textbf{23.51 $\pm$ 0.92} & \textbf{12.17 $\pm$ 1.36} \\
& \textcolor{red}{(+9.50)} & \textcolor{blue}{(-11.86)} \\
\bottomrule
\end{tabular}%
}
\vspace{0.1cm}
\footnotesize{$^{\dagger}$Results from MFP3D~\cite{ma2024mfp3d}}\\
\end{table}

Volume estimation yields a 23.51\% MAPE, compared to MFP3D's 16.15\%. This asymmetry occurs because energy depends on learned food composition features that transfer across datasets, while volume requires absolute spatial measurements tied to the camera parameters of a dataset. 

\subsection{Ablation Studies}
\label{sec:ablation}

We conduct ablation studies on MetaFood3D using a single seed with validation-optimized hyperparameters. 

\subsubsection{Baseline and Adapter Comparison}

\begin{table}[h]
\centering
\caption{Baseline and adapter ablation on MetaFood3D}
\label{tab:baseline_adapter}
\resizebox{\columnwidth}{!}{%
\begin{tabular}{lcc}
\toprule
\textbf{Configuration} & \textbf{Vol. MAE (mL)}~\textcolor{blue}{$\downarrow$} & \textbf{Eng. MAE (kcal)}~\textcolor{blue}{$\downarrow$}\\
\midrule

Baseline (no adapter//distill)$^*$ & 29.18 & 36.37 \\
PortionNet (RGB) & 21.74 & \textbf{29.03} \\
PortionNet (RGB+PC) & \textbf{21.22} & 29.13 \\
\midrule

No Adapter (direct RGB)$^\dagger$ & 27.92 & 38.66 \\
With Adapter (PortionNet) & 26.97 & 33.70 \\
\bottomrule
\end{tabular}%
}
\small
$^*$Dual encoders + PointNet with concatenation. \\
$^\dagger$Trained without distillation, inference bypasses the adapter.
\end{table}

We use a baseline without the adapter or distillation. Table~\ref{tab:baseline_adapter} shows PortionNet achieves 25.5\% volume and 20.2\% energy improvement. Removing distillation ($\lambda_{\text{distill}}=0$) causes significant degradation: 33.2\% and 28.4\% increases in energy and volume MAE, confirming that the adapter requires explicit distillation supervision for effective learning.



\subsubsection{Hyperparameter Optimization}

\begin{table}[h]
\centering
\caption{Hyperparameter ablation on MetaFood3D}
\label{tab:hyperparams}
\resizebox{\columnwidth}{!}{%
\begin{tabular}{lcc}
\toprule
\textbf{Configuration} & \textbf{Vol. MAE (mL)}~\textcolor{blue}{$\downarrow$} & \textbf{Eng. MAE (kcal)}~\textcolor{blue}{$\downarrow$} \\
\midrule
\multicolumn{3}{l}{\textit{RGB-only Training Ratio ($\alpha$), $\lambda{=}0.5$}} \\
$\alpha{=}0.0$ & 27.57 & 34.82 \\
$\alpha{=}0.3$ & \textbf{21.74} & \textbf{29.03} \\
$\alpha{=}0.5$ & 27.13 & 39.69 \\
\midrule
\multicolumn{3}{l}{\textit{Distillation Weight ($\lambda$), $\alpha{=}0.3$}} \\
$\lambda{=}0.0$ & 37.07 & 37.00 \\
$\lambda{=}0.5$ & \textbf{21.74} & \textbf{29.03} \\
$\lambda{=}1.0$ & 23.62 & 36.32 \\
\bottomrule
\end{tabular}%
}
\small
RGB-only inference. 
\end{table}

We validate two critical hyperparameters (Table~\ref{tab:hyperparams}). The 
\textbf{RGB-only training ratio} ($\alpha$) determines the proportion of batches using adapter-generated versus real point cloud features. Insufficient ($\alpha=0$) or excessive ($\alpha=0.5$) ratios degrade performance, while optimal $\alpha=0.3$ achieves 21\% error reduction. The \textbf{distillation weight} ($\lambda$) guides the feature alignment strength between the adapter and PointNet. Ablating distillation ($\lambda=0$) causes 70\% and 28\% degradation in volume and energy, while optimal $\lambda=0.5$ demonstrates effective transfer with 41\% and 22\% improvements. Excessive weighting ($\lambda=1.0$) over-prioritizes feature matching, degrading the performance.
\section{Conclusion}
\label{sec:conclusion}

We presented PortionNet, a novel cross-modal knowledge distillation framework that enables accurate nutrition estimation from RGB images without depth sensors. By training a lightweight adapter to mimic point cloud geometric features, our approach achieves significant improvements over prior methods while requiring only a single image at inference.

Our results show that cross-modal distillation can effectively transfer geometric knowledge from 3D to 2D representations, enabling nutrition estimation on standard devices without specialized hardware.

For future work, one promising direction is to address the volume estimation domain shift across imaging devices by developing camera-agnostic geometric representations through computational photography techniques which could provide consistent scale estimation independent of camera parameters.

{
    \small
    \bibliographystyle{ieeenat_fullname}
    \bibliography{main}
}

\end{document}